\documentclass{article}

\usepackage{arxiv}

\usepackage[utf8]{inputenc} 
\usepackage[T1]{fontenc}    
\usepackage{hyperref}       
\usepackage{url}            
\usepackage{booktabs}       
\usepackage{amsfonts}       
\usepackage{nicefrac}       
\usepackage{microtype}      
\usepackage{lipsum}        
\usepackage{graphicx,epstopdf}
\usepackage{amsmath}
\usepackage{array}

\title{Water Preservation in Soan River Basin Using Deep Learning Techniques}

\author{
  Sadaqat ur Rehman \\
  Department of Electronic Engineering\\
  Tsinghua University\\
  Beijing, 10084 \\
  \texttt{z-sun15@mails.tsinghua.edu.cn} \\
   \And
 Zhongliang Yang  \\
  Department of Electronic Engineering\\
  Tsinghua University\\
  Beijing, 10084 \\
  \texttt{yangzl15@mails.tsinghua.edu.cn} \\
   \And
 Muhammad Shahid \\
  State Key Laboratory of Hydroscience and Engineering\\
  Tsinghua University\\
  Beijing, 10084 \\
  \And
 Nan Wei \\
  National Key Laboratory for Software Technology\\
  Nanjing University\\
  Nanjing, 210023 \\
   \And
 Yongfeng Huang\\
  Department of Electronic Engineering\\
  Tsinghua University\\
  Beijing, 10084 \\
  \texttt{yfhuang@tsinghua.edu.cn} \\
   \And
 Muhammad Waqas \\
  Department of Electronic Engineering\\
  Tsinghua University\\
  Beijing, 10084 \\
  \texttt{wa-j15@mails.tsinghua.edu.cn} \\
   \And
 Shanshan Tu \thanks{Corresponding author.}\\
 Beijing Key Laboratory of Trusted Computing\\
  Beijing University of Technology,\\
  Beijing, 100124 \\
  \texttt{sstu@bjut.edu.cn} \\
   \And
 Obaid ur Rehman \\
  Department of Electrical Engineering\\
  Sarhad University of Science and IT\\
  Pakistan, 25000 \\
  \texttt{obaid.ee@suit.edu.pk} \\
}

\begin{document}
\maketitle

\begin{abstract}
Water supplies are crucial for the development of living beings.  However, change in the hydrological process i.e. climate and land usage are the key issues. Sustaining water level and accurate estimating for dynamic conditions is a critical job for hydrologists, but predicting hydrological extremes is an open issue. In this paper, we proposed two deep learning techniques and three machine learning algorithms to predict stream flow, given the present climate conditions. The results showed that the Recurrent Neural Network (RNN) or Long Short-term Memory (LSTM), an artificial neural network based method, outperform other conventional and machine-learning algorithms for predicting stream flow. Furthermore, we analyzed that stream flow is directly affected by precipitation, land usage, and temperature. These indexes are critical, which can be used by hydrologists to identify the potential for stream flow. We make the dataset publicly available \textit{\textbf{(https://github.com/sadaqat007/Dataset)}} so that others should be able to replicate and build upon the results published.
\end{abstract}


\section{Introduction}

\begin{figure*}[h!]
\centering
\includegraphics[height=13cm,width=0.8\linewidth]{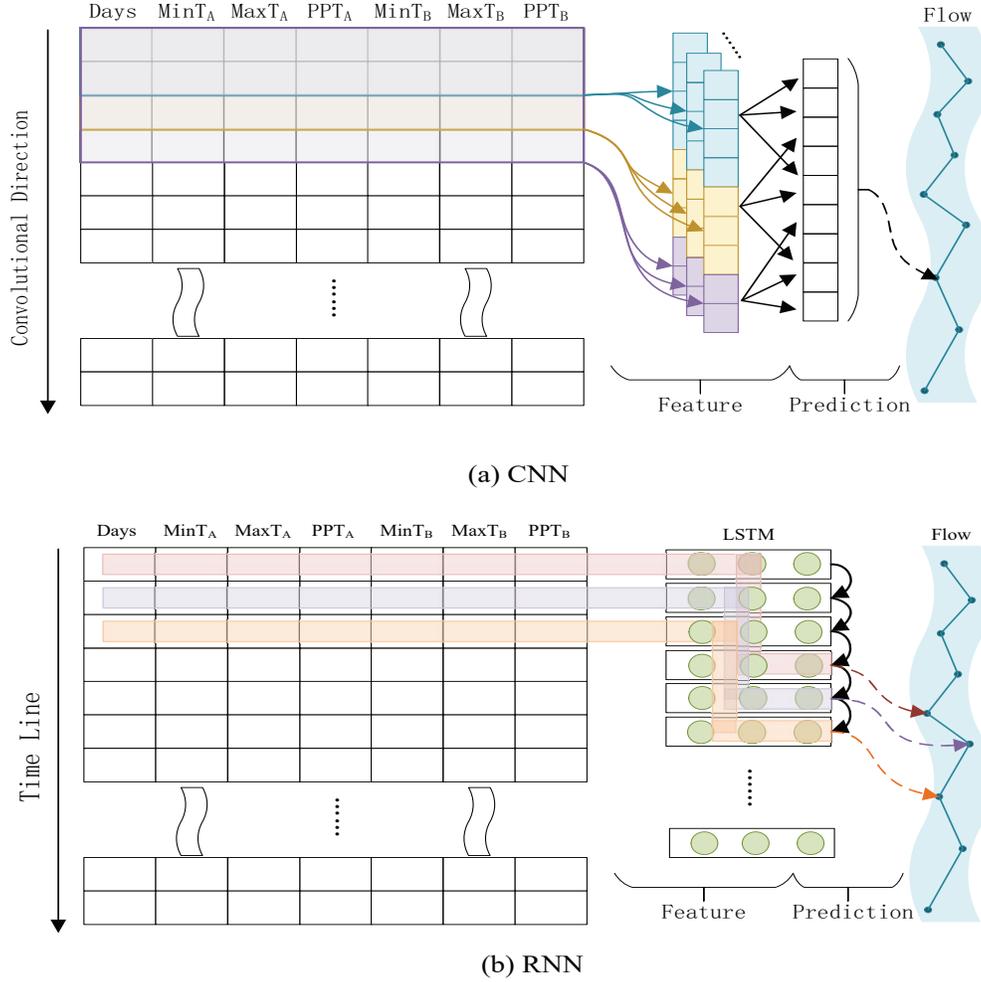}
\caption{The overall framework of the proposed model, (a) CNN, (b) LSTM. These two models are used to extract the feature vectors of amorphous streamflow data and represent them to the feature space, which is further use for predictive analysis}
\label{CNN-LSTM}
\end{figure*}

Water is an important element for all human beings as well as all living things on the earth. Most of the water for domestic, industrial and agriculture use is obtained from the streamflow. The researchers concluded that streamflow has changed across the world \cite{labat2004evidence, gedney2006detection}.

Streamflow is the main component of the hydrological cycle and it is essential for the stability of our ecosystem. Streamflow has a vital impact on the quality of water, living beings, and habitat in the stream \cite{petty2018streamflow}. Streamflow controls the kinds of organisms that can survive in the stream, for example, some organisms are susceptible to fast-flowing areas, while others need stagnant pools. Moreover, it also controls the quantity of residue and sediment carried by the stream. Large, rapidly flowing rivers are less affected by pollution, whereas narrow small streams have less capability to dilute and reduce wastes \cite{zhao2016effects}.

Automating and management of streamflow have the potential to significantly reduce the withdrawals of water for irrigation purposes, as well as to improve the quality of agriculture. The amount of water moving off the watershed into the stream channel is highly related to the flow of a stream. It is affected by the variation in weather, under different conditions such as rainstorm and dry periods. Moreover, different seasons also fluctuate stream flow level, for example, when evaporation is high and the ground is fertile thus removing water from the ground due to which streamflow level decreases and vice versa. Another serious problem of streamflow depletion is water withdrawals for irrigation purposes, for example, industrial water withdrawals \cite{li2017evaluating, naz2018effects}. 

\begin{figure*}[h!]
\centering
    \makebox[\textwidth][l]{\includegraphics[width=17.5cm, height=16cm]{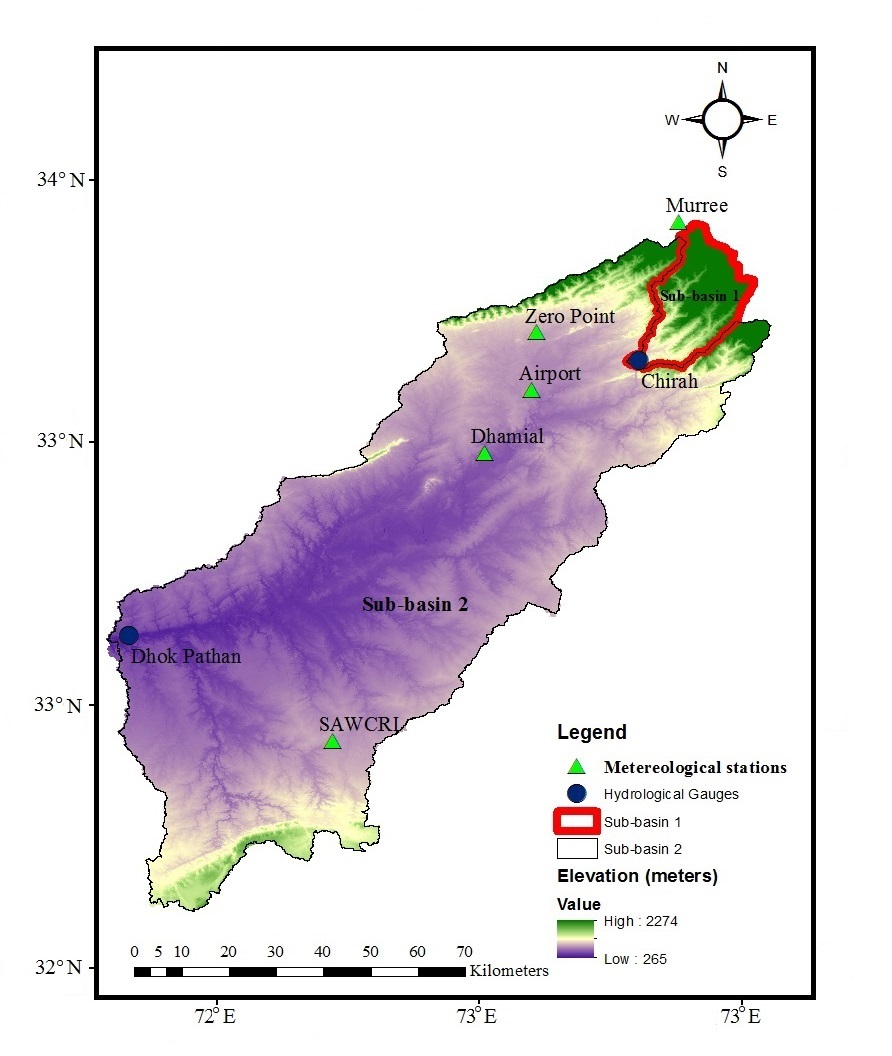}}
    \caption{The Location map of the Soan river basin and Distribution of Hydro-meteorological station}
    \label{fig:Tpserver100}
\end{figure*}

Streamflow estimation and forecasting are challenging and essential for better provision and management of water resources \cite{goyal2013application}. The streamflow prediction can be useful for hydropower generation. Hydrologists have developed different models and methods to estimate the streamflow \cite{mosavi2018flood}. Recently with the development of the computer, programming attempts have been made to estimate streamflow. The artificial neural network (ANN) has been applied successfully for rainfall-runoff modeling, discharge prediction, sediment quantification, and inflow forecasting \cite{hu2018deep, jang2018estimation, liang2018dongting}

In this paper, we applied deep learning techniques for high-level semantic understanding for simulating the daily streamflow. Specifically, two deep learning models: CNN and LSTM are adapted. This work is of great significance and has aroused a wide spread of concern \cite{YZong}. In the current era, deep learning techniques have made significant development in fields such as artificial intelligence \cite{krizhevsky2012imagenet, simonyan2014very, szegedy2015going} and natural language processing \cite{kalchbrenner2014convolutional, kim2014convolutional}. These advancements of deep learning techniques can help the merging of different technologies and inventions in other fields. Many applications are witness that deep learning techniques have benchmark ability in feature extractions and predictions \cite{krizhevsky2012imagenet, zeiler2014visualizing}, which is free from conventional handy feature extraction methods but carries out self-learning through data. This can greatly leverage hydrological extremes, and alleviate the problem of prediction for stream flow. In a nutshell, we don’t need to implement a complex mathematical model for information learning and decision making, but the deep learning models itself can automatically extract useful information from the hydrological condition records by self-learning, and then conduct streamflow prediction based on these information. This makes our model efficient and robust compared to hand-crafted models. The complete overview of the proposed framework is illustrated in Fig. \ref{CNN-LSTM}

Thus, we selected a tropical basin, the Soan River basin, which is the left bank tributary of the Indus River basin, for streamflow assessment. The selected area is very crucial as it has high geographical importance. It is the chief drinking water source to Islamabad city, Pakistan, and the key source of irrigation water for Potohar plateau. In the past few decades, it undergoes a significant climatic and land use \cite{yu2013indus, shahid2014evaluation}. To our best knowledge, we took the first effort to commence the trend analysis of hydrological variables using deep learning algorithms for predicting streamflow change in the Soan river basin. consequently, in this paper, we present and compare results from two different deep learning models, such as CNN and RNN, and some statistical machine learning models to assess the streamflow of the Soan River basin. The main aims of this work are evaluating the performance of different deep learning and machine learning techniques in predicting the stream flow in Soan river basin (Punjab, Pakistan) based on the existing hydrological variables, and to decide an efficient and robust model for this task.


\section{Dataset Evaluation}
\subsection{Soan river basin. }
In this paper, our main focus is on the Soan river basin. The people of Islamabad (capital territory of Pakistan) obtain $95\%$ drinking water from a simly dam, which is the chief reservoir for the Soan river basin. Soan river initiating from Murree Mountains and runs through the Chirah hydrological station, a major branch of Indus River and a significant hydrological unit of Potohar plateau Pakistan. Figure~{\ref{fig:Tpserver100}} shows the visual description of the whole site. After that, it is joined by different branches and at Dhoke Pathan hydrological station it falls into the Indus River. Sub-basin1 is the area above the Chirah hydrological station called Chirah sub-basin, and sub-basin 2 is the area above Dhoke Pathan hydrological station called Dhoke Pathan sub-basin. The Soan basin covered area is $6842 km^2$ having elevation in the range of  $265m - 2274m$. Statistical data obtained from Pak govt. for the period of $1983-2012$ shows that the annual precipitation of the basin is $1465mm$ and the mean annual temperature range from $8^{\circ}C$ to $18^{\circ}C$. The Soan basin has both gentle and steep slope, stream flow in the basin mostly comes from monsoon rainfall. About $60\%$ of the population associated is rural and cultivation is the main source of economy. During last few years, the population is greatly affected by the unpredictable variation in stream flow of soan river basin, therefore, the implementation of a highly predictable performance learning system is the need of the day to assess the variation of streamflow for agriculture purpose \cite{ashfaq2014spatial, shahid2018understanding}.




\subsection{Data description.}
Pakistan meteorological department (PMD) and Soil and Water Conservation Research Institute (SAWCRI) provides the climatological data for this work. The key hydroclimatic parameters, such as temperature and precipitation, were obtained from six different meteorological stations. Water and Power Development Authority (WAPDA) and Capital Development Authority (CDA) provide the daily discharge data of two hydrological stations. Mwangi \cite{mwangi2016relative} et al., elaborate the discharge of Soan basin in detail and computed the runoff for the basin, sub-basin1 and sub-basin2 using a conventional method. We obtain statistical data for thirty years (1983-2012). The survey of Pakistan provides the topographic information of the study area. Punjab bureau of statistics and the Department of Agriculture (Marketing Wing) provided the agriculture production information. The complete data descriptions of hydro-meteorological variables are summarized in Table \ref{tab:summary}.

In this work, the predictors are precipitation and temperature, whereas streamflow level is the response variable. These parameters are selected based on a wealth of case studies. Specifically, streamflow level is highly sensitive to precipitation. For similar reasons, as in Ficchi \cite{ficchi2014short} et al., we also include only two meteorological variables, such as precipitation and temperature to provide a challenging scenario for learning algorithms to leverage streamflow. 

\begin{table}[h]
\renewcommand\arraystretch{1.5}
\centering
\caption{\label{tab:2} Hydrological Characteristics of the Soan basin and its sub-basins, 1983-2012.}
\begin{tabular}{| m{2cm} |m{2cm}| m{2cm} | m{3cm} |m{2cm}| m{2cm} |}
\toprule[1.5pt]
Area of Study &Area/km2 &Precipitation (mm/year) & Potential Evapotranspiration (mm/year)&Runoff (mm/year)&Evaporation (mm/year)\\
\hline
Soan Basin &6675 &972.1 &1382.5&241.8&730.3\\
\hline
Sub-basin 1 &335 &1649.4&934.5&695.8&953.6\\
\hline
Sub-basin 2 &6340 &988.3&1394.1& 190.5&797.8\\
\hline
\end{tabular}
\label{tab:summary}
\end{table}

\section{Models Design and Analysis.}
In this study, we propose two different learning algorithms to extract features from hydroclimatic variables and predict streamflow levels. Specifically, CNN and RNN. As we discussed before, the key aim of selecting these different types of learning algorithms is to validate streamflow prediction on a variety of algorithms. The overall system frameworks are shown in Figure \ref{CNN-LSTM}. Detail descriptions and mathematical models for the individual algorithm are as follow.

\subsection{CNN-based prediction model.}

Convolutional Neural Networks are the class of deep learning and is very similar to ordinary Neural Networks. They are made up of neurons that have learnable weights and biases. Each neuron receives some inputs, performs a dot product and optionally follows it with a non-linearity. The whole network expresses a single differentiable score function from the raw image pixels on one end to class scores at the other \cite{rehman2018optimization, tu2017csfl}. For the CNN-based prediction model, a convolution neural network is introduced to extract features. The input to the model is the daily temperature and rainfall information for the upstream cities, and the output is the predicted river flow. The architecture of the proposed CNN for streamflow prediction is as follows: 

It consists of a convolutional layer with three various sizes of convolutional kernels, followed by an average pooling layer and a fully-connected (fc) layer with a supervision signal called softmax loss. 
 
The function of the convolutional layer is to extract features map from the input data matrix and different sizes convolutional kernels can learn various low-level features. The pooling layer sufficed as a downsampling layer to boost up system efficiency and robustness  \cite{krizhevsky2012imagenet, boureau2010theoretical}. Moreover, all these features are fused in the fully-connected layer, which further passes them to a supervision signal called softmax classifier for streamflow prediction. The purpose of the softmax classifier is to compute the similarity measurement between input feature vectors and finally display the streamflow probability. Detail description of parameter tuning is given in the ``Experimental Setting" section. 

For individual input data matrix, we describe it with a matrix $X \in \mathbb{R}^{N \times D}$, as illustrated in Eq. \ref{eq-1}. Whereas the $i$-th row suggests the $i$-th day, and $D$ represents the dimension of the input feature. Mathematically it can be represented as:

\begin{equation}   
X = 
\left[                 
  \begin{array}{c}   
    x_{1}\\  
    x_{2}\\  
    \vdots\\
    x_{N}\\
  \end{array}
\right]
=
\left[                 
  \begin{array}{cccc}   
    a_{1,1} & a_{1,2} & \cdots\ & a_{1,D}\\  
    a_{2,1} & a_{2,2} & \cdots\ & a_{2,D}\\  
    \vdots & \vdots & \ddots & \vdots\\
    a_{N,1} & a_{N,2} & \cdots\ & a_{N,D}\\
  \end{array}
\right]                 
\label{eq-1}
\end{equation}

Generally, let $X_{i:j}$ refer to the matrix which feature vectors from the $i$-th day to the $j$-th day, mathematically:

\begin{equation}    
X_{i:j} = 
\left[                 
  \begin{array}{c}   
    x_{i}\\  
    x_{i+1}\\  
    \vdots\\
    x_{j}\\
  \end{array}
\right]
=
\left[                 
  \begin{array}{cccc}   
    a_{i,1} & a_{i,2} & \cdots\ & a_{i,D}\\  
    a_{i+1,1} & a_{i+1,2} & \cdots\ & a_{i+1,D}\\  
    \vdots & \vdots & \ddots & \vdots\\
    a_{j,1} & a_{j,2} & \cdots\ & a_{j,D}\\
  \end{array}
\right]                 
\end{equation}

Each convolution layer is made up of several convolution kernels of various sizes, and every size further contains many convolution kernels. The width of the convolution kernel is aligned with the width of the input data matrix. For example, if the height of $k$-th convolution kernel is $H$, then it can be described as  $W^k \in \mathbb{R}^{H \times D}$, mathematically:

\begin{equation}    
W^k = 
\left[                 
  \begin{array}{cccc}   
    w^k_{1,1} & w^k_{1,2} & \cdots\ & w^k_{1,D}\\  
    w^k_{2,1} & w^k_{2,2} & \cdots\ & w^k_{2,D}\\  
    \vdots & \vdots & \ddots & \vdots\\
    w^k_{H,1} & w^k_{H,2} & \cdots\ & w^k_{H,D}\\
  \end{array}
\right]             
\end{equation}

The function of convolution operation is the features extraction for local region elements of the input data matrix. For instance, when $w^k_{1,1}$ and $a_{1,1}$ co-occur, then the feature $c^k_1$ extracted from $X_{1:H}$ by the convolutional kernel can be expressed as:
\begin{equation} 
c^k_1 = f(\sum_{i=1}^{H}\sum_{j=1}^{D} w^k_{i,j}\cdot a_{i,j} + b^k_{i,j}), 
\end{equation}
\noindent weight $w^k_{i,j}$ represents the significance of the $j$-th value in the $i$-th row vector, $b^k_{i,j}$ denote the bias term and $f$ is a nonlinear function, as described in \cite{krizhevsky2012imagenet}. The $ReLu$ function is described as nonlinear function, which is mathematically illutrated as:
\begin{equation} 
z=ReLu(i) = max(0,i). 
\end{equation} 
The convolution process is obtained by doing the convolution between the kernels $W^k$ and input data matrix $X$ with a certain step of  $T_c$, and computes the features of individual local region. In conclusion, the feature extracted by the convolution kernel $W^k$ can be expressed as:

\begin{equation} 
C^k = [c^k_1,c^k_2,...,c^k_{\frac{N-H+1}{T_c}}]^\top. 
\end{equation} 
The pooling layer served to lessen the amount of neural network parameter and maintain the overall data distribution, which eventually prevent the system from over-fitting problem and enhance system robustness \cite{krizhevsky2012imagenet, boureau2010theoretical}. Both the pooling and convolution operations have similar function, with a difference that pooling operatikon computes the maximum or average value of the local area. Here, after obtaining the feature $C^k$ from convolution operation we apply a max pooling operation. Let, $H_p$ represent the height of a pooling kernel and $T_p$ represent its step size, then mathematically we compute the output as:

\begin{equation} 
M^k = [m^k_1,m^k_2,...,m^k_{N_p}]^\top, 
\end{equation} 

\noindent where

\begin{equation} 
m^k_i = max(c^k_{i},c^k_{i+1},...,c^k_{i+H_p-1}). 
\end{equation} 

\begin{equation} 
N_p = \frac{\frac{N-H+1}{T_c}-H_p+1}{T_p}, 
\end{equation} 

The aforementioned process described that a single convolution kernel $W^k$ produces single feature map $M^k$. Finally, after the successful completion of convolution and pooling operations, the entire extracted features are combined in an end-to-end fashion to achieve a single feature vector of the whole input data matrix, which can be mathematically expressed as: 

\begin{equation} 
F^\top = [F^\top_1;F^\top_2;...;F^\top_l], 
\end{equation} 
\noindent where $F_i = M^i$, $l$ represents the total number of features.

Finally, for high-level features extraction fully-connected layer is used. Let, $W_F$ represent the weight matrix, in order to calculate the weighted sum of the individual element to get the final feature representation of the input data matrix $S$, mathematically:

\begin{equation}
y = W_F \cdot F + b_f,
\end{equation}

\noindent where $W_F$ represents the learned weight matrix and $b_f$ represents the bias term, however, the weight matrix $W_F$ highlights the significance of every individual feature.

The parameters of the network are updated using backpropagation algorithm in the training process, and the loss function is computed through Smooth-L1 loss called Huber loss:
\begin{equation}
loss(\hat{y}, y) = \frac{1}{n}\sum_i z_i,
\end{equation}
where $z_i$ can be calculated as:
\begin{equation}
z_i =
\begin{cases}
0.5(\hat{y}-y)^2, & \text{if}~ |\hat{y}-y| < 1\\
|\hat{y}-y|-0.5, & \text{otherwise}.
\end{cases}
\end{equation}

The main choice of selecting Smooth-$L_1$ loss is to diminish data sensibility, as outliers greatly affect $L_2$ loss. In this paper, we used the minimization of the $LOSS$ function over a batch size number of samples for CNN training. The optimization algorithm is \textsc{Ada-delta} with $1.0$ initial learning rate. 

\subsection{LSTM-based prediction model}

Recurrent Neural Network (RNN) is an artificial neural network architecture, which is very suitable for sequential signals modeling. Mathematically, we described the RNN model in Eq. $(8)$, which can deal with randomly size sequential signals. Though, the system is not able to handle long-range problem efficiently due to the gradient vanishing problem \cite{hochreiter1998vanishing}. LSTM model \cite{Hochreiter1997Long}, a variant of RNN efficiently solve this problem through rich designed unit nodes. The key architecture refinement of LSTM is the hidden layer unit, which contained four fundamental parts: a cell, an input gate, an output gate and a forget gate. The basic LSTRM mathematical model is illustrated as follows: 

\begin{flalign}
& \left\{\begin{array}{l}
 I_t = \sigma(W_i \cdot [h_{t-1},x_t] + b_i), \\
 F_t = \sigma(W_f \cdot [h_{t-1},x_t] + b_f),\\
 C_t = F_t \cdot C_{t-1} + I_t \cdot \tanh(W_c \cdot [h_{t-1},x_t] + b_c),\\
 O_t = \sigma(W_o \cdot [h_{t-1},x_t] + b_o),\\
 h_t = O_t \cdot \tanh(C_t).\\
             \end{array}  
        \right.&
\end{flalign}

\noindent where $I_t$ represents the input gate, which controls the quantity of novel information stored in the memory cell. The function of forget gate $F_t$ is to discard redundant stored information from the memory cell. Hence, memory cell $C_t$ is a summation of upcoming and previous information modulated by the input gate and the forget gate, respectively. The output gate $O_t$ allows the memory cell to have an effect on the current hidden state and output or block its influence. Specifically, we represent the transfer function of LSTM units by $f_{LSTM}(\ast)$. Notice that, while calculating the output at time step $t$, the information used is based on the input vector at time step $t$, also includes the information stored in the cells at the previous time step $t-1$. Hence, the output at a given time steps $t$ can be described as: 

\begin{equation}
y_t = f_{LSTM}(x_t\mid x_1,x_2,...,x_{t-1}).
\end{equation}

As we have mentioned before, the input matrix $X$ can be regarded as a sequential signal and the $i$-th row $x_i$ can be viewed as the signal at time step $i$. In order to model this task on LSTM, we input the $i$-th row of $X$ at time step $i$. For individual sentence $S$, we can describe it with a matrix $S \in \mathbb{R}^{N \times D}$ as described in Eq. \ref{eq-1}.

Overall, an RNN contains several network layers, each of which with multiple LSTM units. Let, $n_j$ represent the number of LSTM units  $U^j$ of $j$-th hidden layer, thus the units of $j$-th layer can be described as:
\begin{equation}
U^j = \{u^j_{1},u^j_{2},\cdots,u^j_{n_j}\}.
\end{equation}
In LSTM first hidden layer, the input of each unit $u^1_{i}$ at time step $t$ is the weighted sum of the elements in $x_t$, mathematically as:

\begin{equation}
u^1_{i,t} = \sum^d_{k=1}w^1_{i,k}\cdot a_{j,k} + b^1_{i,t},
\end{equation}

\noindent where $w^1_.$ represents the learned weights and $b^1.$ represents the biases. Hence, the output value of $u^1_{i}$ at time step $t$ is:

\begin{equation}
o^1_{i,t} = f_{LSTM}(u^1_{i,t}) = f_{LSTM}(\sum_{k=1}^dw^1_{i,k}\cdot a_{j,k} + b^1_{i,t}).
\end{equation}

The vector $Out^j_t$ is used to represent the output of the $j$-th hidden layer at time step $t$, and each element in $Out^j_t$ represents the output value of each unit in the $j$-th hidden layer at time step $t$, mathematically:

\begin{equation}
Out^j_t = f_{LSTM}(x_t) = [o^j_{1,t},o^j_{2,t},...,o^j_{n_j,t}].
\end{equation}

Literature showed that increasing the number of neural network layers, increasing the ability of system to extract features \cite{Zeiler2013Visualizing}. Therefore, in this paper, the network is stack with multiple layers of LSTM units. The transfer matrix is used to connect the adjacent hidden layers. For instance, the transfer matrix between $l$-th layer and $(l+1)$-th layer can be described as a matrix $W^{l} \in \mathbb{R}^{{n_l} \times {n_{l+1}}}$:

\begin{equation}
W^{l} = \left[                 
  \begin{array}{cccc}   
    w^l_{1,1} & w^l_{1,2} & \cdots\ & w^l_{1,n_{l+1}}\\  
    w^l_{2,1} & w^l_{2,2} & \cdots\ & w^l_{2,n_{l+1}}\\  
    \vdots & \vdots & \ddots & \vdots\\
    w^l_{n_l,1} & w^l_{n_l,2} & \cdots\ & w^l_{n_l,n_{l+1}}\\
  \end{array}
\right]
\end{equation}

Then the input of each unit $u^l_{i}$ in the $l$-th hidden layer at time step $t$ is the weighted sum of the output values of the units in the previous layer, mathematically: 

\begin{equation}
u^l_{i,t} = Out^{l-1}_t \cdot W^{l,m} = \sum_{k=1}^{n_{l-1}}w^l_{i,k}\cdot o^{l-1}_{k,t} + b^l_{i,t}.
\end{equation}

Furthermore, the output of the $l$-th layer at time interval $t$ is:

\begin{equation}
\begin{aligned}
&Out^l_t = f_{LSTM}(Out^l_t) = [o^l_{1,t},o^l_{2,t},...,o^l_{n_l,t}],\\
&o^l_{i,t} = f_{LSTM}(u^l_{i,t}) = f_{LSTM}(\sum_{k=1}^{n_{l+1}}w^{l}_{i,k}\cdot o^l_{k,t} + b^{l}_{i,t}).
\label{equation22}
\end{aligned}
\end{equation}

The aforementioned Eq. \ref{equation22} illustrated that the output at time step $t$ is not only based on the input vector of the current time $x_t$, but also the information stored in the cells at the previous $(t-1)$ intervals. Thus, $l$-th hidden layer output at time interval $t$ can be obtained as a summation of all former $t$ intervals, such as information fusion of former $t$ intervals $\{x_1,x_2,...,x_t\}$. Moreover, to compute the probability distribution of the $(t+1)$ day, a softmax layer is added after all the hidden layers of the LSTM. Particularly, we introduce the Prediction Weight (PW) as matrix $W_P \in \mathbb{R}^{{n_l} \times N}$, mathematically as:

\begin{equation}
W_P = \left[                 
  \begin{array}{cccc}   
    w^p_{1,1} & w^p_{1,2} & \cdots\ & w^p_{1,N}\\  
    w^p_{2,1} & w^p_{2,2} & \cdots\ & w^p_{2,N}\\  
    \vdots & \vdots & \ddots & \vdots\\
    w^p_{n_3,1} & w^p_{n_3,2} & \cdots\ & w^p_{n_l,N}\\
  \end{array}
\right]
\end{equation}

\noindent where $N$ represents the total number of days in the dictionary $D$. Then the learned matrix $W_P$ is used to compute  the score for individual day in the dictionary $D$ for a time span of 1983-2012. Mathematically, we can define as:

\begin{equation}
y_i = \sum^{n_l}_{k=1}w^p_{k,i}\cdot o^l_{i,t} + b^p_{i,t},
\end{equation}

\noindent where $W_P$ represents the learned weight and $b^p$ represents the bias term, the values in weight matrix $W_P$ reflect the significance of individual feature in $o^l$. Then we can describe the dimension of output vector $y$ as $N$. 

Both CNN and LSTM based prediction models use the same Loss function and training algorithm.

\section{Experimental Results and Discussion}

\begin{figure*}
\centering
\includegraphics[height=5cm,width=0.5\linewidth]{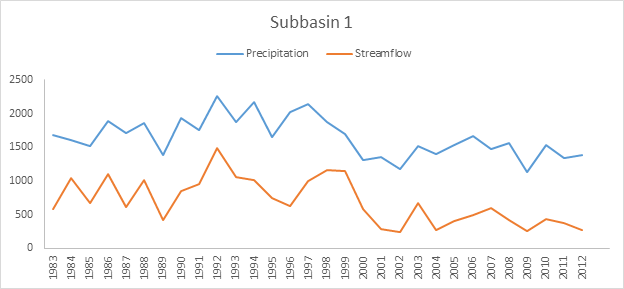}
\caption{Precipitation and stream flow statistics of the Soan sub-basin1 for the year 1983-2012}
\label{subbasin-1}
\end{figure*}

\begin{figure*}
\centering
\includegraphics[height=5cm,width=0.5\linewidth]{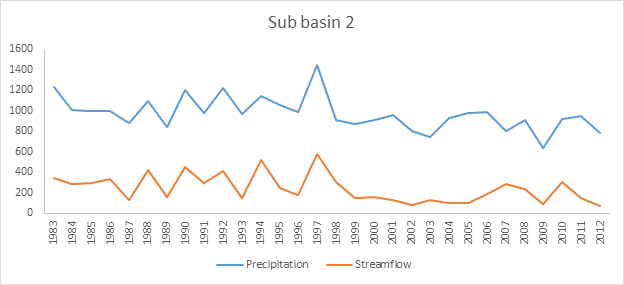}
\caption{Precipitation and streamflow statistics of the Soan sub-basin2 for the year1983-2012}
\label{subbasin-2}
\end{figure*}

\subsection{Experiment Setting}

In this paper, we performed all the experiments on two deep learning models, such as CNN and LSTM, and three machine learning algorithms, such as RF, GBR, and LR. For the CNN model, the input data is fed to a convolution layer with three different kernel height: $3, 5, 7$. The width of kernels is equal to the feature number. Each kernel initially generates 100 channels. After the convolution operation, max pooling is applied to each channel. Then the outputs are concatenated together to the last layer, which is a fully-connected layer with a dimension of $300\times1$. The output of which is the predicted flow.

For the LSTM model, we use a bidirectional LSTM with cell number equal to the features number and 300 hidden sizes. Moreover, the max-pooling layer and fully-connected layer follow the same architecture as CNN. 

As for the traditional machine learning methods, such as Logistic Regression (LR) \cite{suykens1999least, harrell2015ordinal}, Gradient Boosting Tree (GBR) \cite{friedman2002stochastic} and Random Forest (RF) \cite{liaw2002classification}, we used the default setting from package \textit{scikit-learn}.

\subsection{Dataset Split}
We divided the dataset into three subsets, such as training set, validation set, and testing set, with a proportion of $7:1:2$. The final result is obtained using a 10 fold cross validation with different random dataset division. Normalization is applied to each feature in the preprocessing. 

Mini-batch is used in the training process of CNN and LSTM, with 90 batch size. The optimization algorithm is \textsc{Ada-delta} with $1.0$ initial learning rate.

\subsection{Evaluation Method}

The index we adapt to evaluate results is the relative error, which is defined as:
\begin{equation}
re = \frac{1}{n}\sum_i\frac{|predict - real|}{real}.
\end{equation}
Notice that we have normalized the raw data in preprocessing, so before calculating relative error, the output will be recovered to the initial distribution.

\subsection{Hydro-climatic Trends in Soan Basin} 
Fig. \ref{subbasin-1}, \ref{subbasin-2} shows the hydro-climatic trends in sub-basin 1 and sub-basin 2 (all units in mm). It can be observed that precipitation and streamflow of both sub-basin have a strong correlation. The catchment area of the sub-basin 1 is smaller as compared to the sub-basin 2, therefore the streamflow of the sub-basin 1 is more as compared to sub-basin 2. The sub-basin 1 has hilly areas and due to high mountains and slope differences, most of the precipitation is converted into streamflow and also has small water loses. On the other hand, sub-basin 2 has less precipitation as compared to sub-basin 1, which eventually causes less streamflow in sub-basin 2. There is more water loss in sub-basin 2 and has high agriculture and water consumption. Moreover, the water storages and small check dams in sub-basin 2 can cause a reduction in the streamflow. While, on the other hand, sub-basin 1 has no small check dams or retention, which can stop the water and topography is favorable for high streamflow and less water loss.

\subsection{Predictive performance of different models.} Supervised learning algorithms are the major benchmark paradigms found in many applications, such as classification, detection and hydrological modeling \cite{nateghi2011comparison, nateghi2016statistical, grizzetti2017human, nishina2017varying}. Supervised learning algorithms contain three fundamental parts: an input object (vector), a processing unit (hidden layers) and an output value (supervision signal). These learning algorithms produce an inferred function by evaluating the training data, which can be further employed for representing new examples. An optimal setting allows these algorithms to describe the class labels for hidden examples, which help the learning algorithms to generalize for unseen conditions from the training data. Let, given a set of $\aleph$ training examples $(x_{i},y_{j}),...,(x_{N})$ such that $x_{i}$ represent the feature vector of $i^{th}$ example and $y_{j}$ is its class label, a supervised learning algorithm search for a function $g:X\to Y$, where $X$ and $Y$ is the input and output space respectively. The function $g$ is an element of the hypothesis space of possible functions $\mathfrak{G}$. Under certain circumstances $g$ is represented as a scoring function $f:X\times Y\to {\mathbb {R}}$, such that, it return the value of $y$ that gives the maximum score: $g(x)={\underset {y}{\arg \max }}f(x,y)$ . Suppose, $F$ represents the scoring functions.

\begin{table}[h]
\renewcommand\arraystretch{1.5}
\centering
\caption{\label{tab:2}Relative Error of different Deep learning and Machine learning methods.}
\begin{tabular}{|l|c|c|}
\toprule[1.5pt]
Model &Mean Relative Error(\%) &Standard Deviation\\
\hline
LR &28.0377 &1.9769\\
\hline
GBR &27.4681 &1.8108\\
\hline
RF &31.5304 &1.9430\\
\hline
CNN &26.8145 &1.6070\\
\hline
LSTM &\textbf{25.0984} &1.9182\\
\bottomrule[1.5pt] 
\end{tabular}
\label{tab:results}
\end{table}

\begin{figure}

\begin{minipage}[H]{0.5\linewidth}
\centering
\includegraphics[width=3.2in]{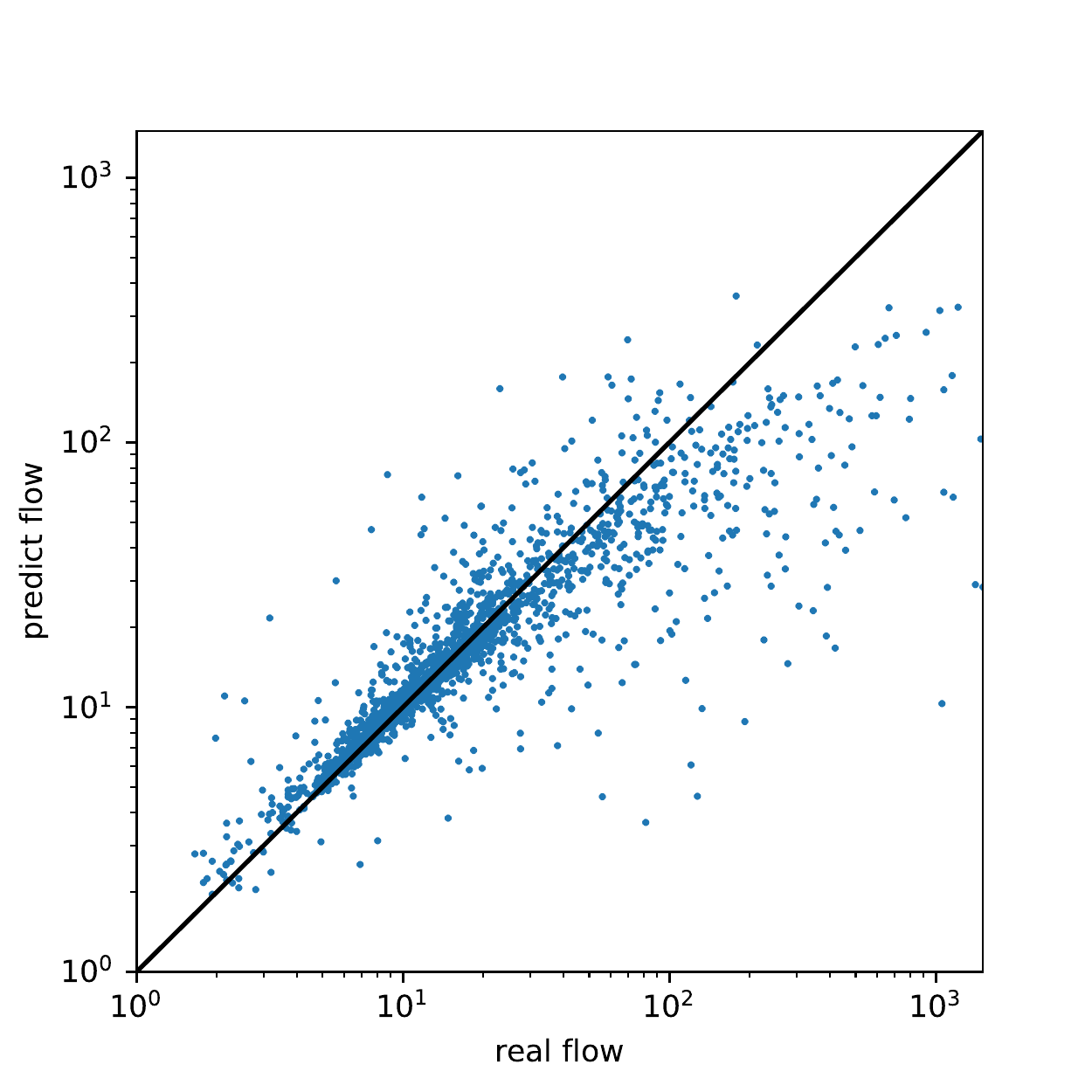}
\caption{Prediction results using LSTM. The horizontal axis shows the real stream flow data, whereas the vertical axis shows the predicted results obtained through LSTM}
\label{fig:side:b}
\end{minipage}
\begin{minipage}[H]{0.5\linewidth}
\centering
\includegraphics[width=3.2in]{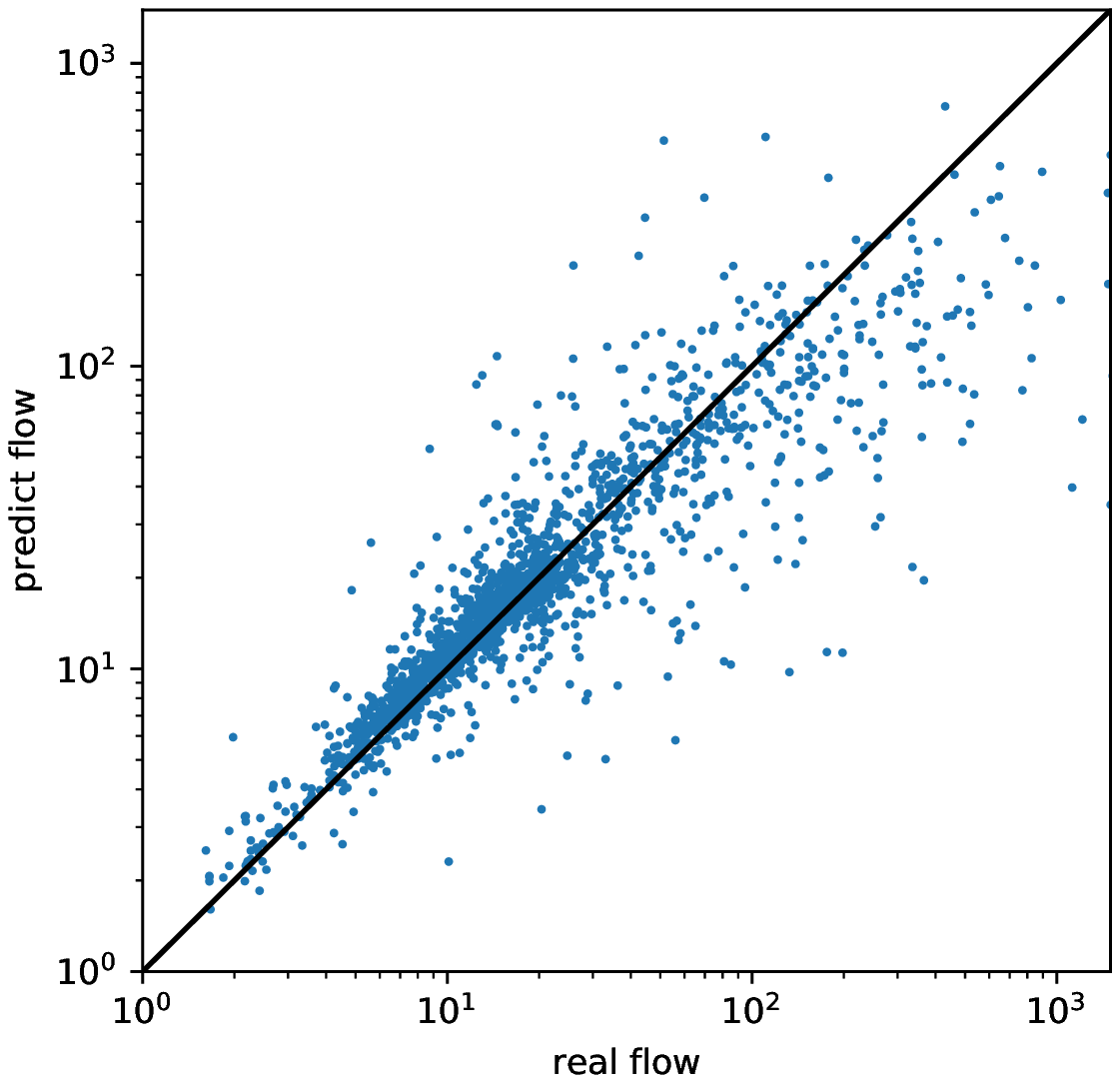}
\caption{Prediction results using CNN. The horizontal axis shows the real streamflow data, whereas the vertical axis shows the predicted results obtained through CNN}
\label{fig:side:a}
\end{minipage}%
\end{figure}

\begin{figure}
\centering
\includegraphics[width=0.6\linewidth]{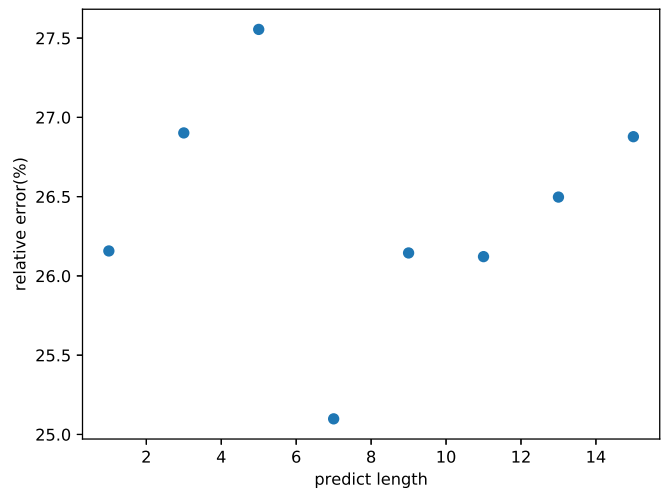}
\caption{Prediction error changes with days}
\label{fig:3}
\end{figure}

Even though, $G$ and $F$ can be any space of functions, many learning algorithms such as CNN and LSTM are probabilistic models where $g$ takes the form of a joint probability model $g(x,y)=P(x,y)$, and logistic regression is a conditional probability model where $f$ takes the form of a conditional probability model $g(x)=P(y|x)$.

In this paper, we evaluated five different learning models for measuring predictive performance – CNN, LSTM, RF, GBR, and LR - to predict the stream flow based on hydrological variables. These different learning models were chosen to test the performance of a wide variety of learning algorithms on streamflow prediction problem. The inclusion of both statistical models, such as LR, GBR, and RF, and deep learning models, such as LSTM and CNN, help to ensure that the predictive performance is tested on a wide variety of learning algorithms. Statistical models are used in a variety of fields, such as medical disease diagnosis, machine learning and social sciences for prediction. Moreover, these models itself simply model the probability of output in terms of input and do not perform statistical classification. CNN and LSTM models can provide robust predictions, due to the transformations of the input space in the inner layers \cite{lancini1997diagnosing}. The outcome of each algorithm was summarized using Mean Relative error (MRE) of successful prediction (error rate \%), as shown in Table \ref{tab:results}. Moreover, the advantage of deep learning models (CNN and LSTM) over statistical models is that deep learning models continue to improve efficiency as the number of datasets grows. Whereas, statistical models plateau at a certain level of performance when you add more examples and training data to the network.

Table \ref{tab:results} shows the parameter settings and numerical results obtained using different algorithms including the proposed algorithms. It is clear from the results that the LSTM surpasses other benchmark algorithms in solving streamflow prediction. It demonstrates that the Random Forest (RF) approach achieved the highest MRE (31.53\% ) on the streamflow prediction dataset. Followed by Improved Linear Regression (LR) with mean relative error (MRE) of 28.03\%. The Gradient Boosted Regression (GBR) algorithm and CNN have similar MRE of 27.46\% and 26.81\% respectively (the overall difference was no more than 0.6\%). The lowest MSE is achieved by the LSTM method, which is 25.09\%. Furthermore, LSTM also shows high predictive accuracy despite having a small error, as shown in Fig. \ref{fig:side:b}. It shows the real stream flow plotted against the predicted flow by LSTM with a $45{\circ}$ reference line. Followed by CNN, which also shows a good predictive accuracy as shown in Fig. \ref{fig:side:a}. Hence, we concluded that the LSTM algorithm with the lowest MRE achieves the highest prediction rate, which needs less training time to obtain an optimum solution.

The accuracy of forecasting traffic is related to the forecast duration. For example, according to the data of the first five days or the first 10 days or the first month, the accuracy of the next day's traffic forecast is different. Therefore, the abscissa of this graph is how many days before we forecast the next day. The ordinate is the accuracy of the prediction. The main purpose is to find the best forecast duration. Based on the data in this Fig. \ref{fig:3}, we find that when the predicted number of days is 7, the predicted error is the smallest. So our model was finally determined to predict river flow of the next day based on data from the previous 7 days.



\section{Conclusions}

Two deep learning techniques and three machine learning algorithms are proposed to predict stream flow, given the present climate conditions. These different learning models were chosen to evaluate the performance of a wide variety of learning algorithms on streamflow prediction problem. The inclusion of both statistical models, such as LR, GBR, and RF, and deep learning models, such as LSTM and CNN, help to ensure that the predictive performance is tested on a wide variety of learning algorithms. The results showed that the LSTM, an artificial neural network based method, outperforms other conventional and machine-learning algorithms for predicting stream flow.

To our best knowledge, this is the first work to introduce five different deep learning and machine learning algorithms for streamflow prediction problem. Furthermore, we believe that the proposed predictive algorithms serve as a reference tool for researchers and engineers to help the design and implementation of improving evaluation protocols.

\section*{Acknowledgments}This work was partially supported by the National Natural Science Foundation of China (No. 61801008, U1705261, U1405254, U1536115, U1536207),  National Key R\&D Program of China (No. 2018YFB0803600, 2016YFB0801301), Beijing Natural Science Foundation National (No. L172049), Scientific Research Common Program of Beijing Municipal Commission of Education (No. KM201910005025).

\vspace{6pt}

\bibliography{sample}

\begin{thebibliography}{10}

\bibitem{labat2004evidence}
David Labat, Yves Godd{\'e}ris, Jean~Luc Probst, and Jean~Loup Guyot.
\newblock Evidence for global runoff increase related to climate warming.
\newblock {\em Advances in Water Resources}, 27(6):631--642, 2004.

\bibitem{gedney2006detection}
Nicola Gedney, PM~Cox, RA~Betts, O~Boucher, C~Huntingford, and PA~Stott.
\newblock Detection of a direct carbon dioxide effect in continental river
  runoff records.
\newblock {\em Nature}, 439(7078):835, 2006.

\bibitem{petty2018streamflow}
TR~Petty and P~Dhingra.
\newblock Streamflow hydrology estimate using machine learning (shem).
\newblock {\em JAWRA Journal of the American Water Resources Association},
  54(1):55--68, 2018.

\bibitem{zhao2016effects}
Gang Zhao, Huilin Gao, and Lan Cuo.
\newblock Effects of urbanization and climate change on peak flows over the san
  antonio river basin, texas.
\newblock {\em Journal of Hydrometeorology}, 17(9):2371--2389, 2016.

\bibitem{li2017evaluating}
Zhi Li and Jiming Jin.
\newblock Evaluating climate change impacts on streamflow variability based on
  a multisite multivariate gcm downscaling method in the jing river of china.
\newblock {\em Hydrology and Earth System Sciences}, 21(11):5531--5546, 2017.

\bibitem{naz2018effects}
Bibi~S Naz, Shih-Chieh Kao, Moetasim Ashfaq, Huilin Gao, Deeksha Rastogi, and
  Sudershan Gangrade.
\newblock Effects of climate change on streamflow extremes and implications for
  reservoir inflow in the united states.
\newblock {\em Journal of Hydrology}, 556:359--370, 2018.

\bibitem{goyal2013application}
Manish~Kumar Goyal, CSP Ojha, RD~Singh, PK~Swamee, et~al.
\newblock Application of artificial neural network, fuzzy logic and decision
  tree algorithms for modelling of streamflow at kasol in india.
\newblock {\em Water Science and Technology}, 68(12):2521--2526, 2013.

\bibitem{mosavi2018flood}
Amir Mosavi, Pinar Ozturk, Shahab Shamshirband, Hai~Thanh Nguyen, and Kwok-wing
  Chau.
\newblock Flood prediction using machine learning.
\newblock {\em Literature Review, Engineering Applications of Computational
  Fluid Mechanics}, 2018.

\bibitem{hu2018deep}
Caihong Hu, Qiang Wu, Hui Li, Shengqi Jian, Nan Li, and Zhengzheng Lou.
\newblock Deep learning with a long short-term memory networks approach for
  rainfall-runoff simulation.
\newblock {\em Water}, 10(11):1543, 2018.

\bibitem{jang2018estimation}
Dongwoo Jang, Hyoseon Park, and Gyewoon Choi.
\newblock Estimation of leakage ratio using principal component analysis and
  artificial neural network in water distribution systems.
\newblock {\em Sustainability}, 10(3):750, 2018.

\bibitem{liang2018dongting}
Chen Liang, Hongqing Li, Mingjun Lei, and Qingyun Du.
\newblock Dongting lake water level forecast and its relationship with the
  three gorges dam based on a long short-term memory network.
\newblock {\em Water}, 10(10):1389, 2018.

\bibitem{YZong}
Sills J. et~al.
\newblock Nextgen voices: Unique identities[j].
\newblock {\em Science}, 363(6428):702--702, 2019.

\bibitem{krizhevsky2012imagenet}
Alex Krizhevsky, Ilya Sutskever, and Geoffrey~E Hinton.
\newblock Imagenet classification with deep convolutional neural networks.
\newblock In {\em Advances in neural information processing systems}, pages
  1097--1105, 2012.

\bibitem{simonyan2014very}
Karen Simonyan and Andrew Zisserman.
\newblock Very deep convolutional networks for large-scale image recognition.
\newblock {\em arXiv preprint arXiv:1409.1556}, 2014.

\bibitem{szegedy2015going}
Christian Szegedy, Wei Liu, Yangqing Jia, Pierre Sermanet, Scott Reed, Dragomir
  Anguelov, Dumitru Erhan, Vincent Vanhoucke, and Andrew Rabinovich.
\newblock Going deeper with convolutions.
\newblock In {\em Proceedings of the IEEE conference on computer vision and
  pattern recognition}, pages 1--9, 2015.

\bibitem{kalchbrenner2014convolutional}
Nal Kalchbrenner, Edward Grefenstette, and Phil Blunsom.
\newblock A convolutional neural network for modelling sentences.
\newblock {\em arXiv preprint arXiv:1404.2188}, 2014.

\bibitem{kim2014convolutional}
Yoon Kim.
\newblock Convolutional neural networks for sentence classification.
\newblock {\em arXiv preprint arXiv:1408.5882}, 2014.

\bibitem{zeiler2014visualizing}
Matthew~D Zeiler and Rob Fergus.
\newblock Visualizing and understanding convolutional networks.
\newblock In {\em European conference on computer vision}, pages 818--833.
  Springer, 2014.

\bibitem{yu2013indus}
Winston Yu, Yi-Chen Yang, Andre Savitsky, Donald Alford, Casey Brown, James
  Wescoat, Dario Debowicz, and Sherman Robinson.
\newblock {\em The Indus basin of Pakistan: The impacts of climate risks on
  water and agriculture}.
\newblock The World Bank, 2013.

\bibitem{shahid2014evaluation}
Muhammad Shahid, Hamza~Farooq Gabriel, Amjad Nabi, Sajjad Haider, AS~Khan, and
  AMS Shah.
\newblock Evaluation of development and land use change effects on
  rainfall-runoff and runoff-sediment relations of catchment area of simly lake
  pakistan.
\newblock {\em Life Science Journal}, 11(7s), 2014.

\bibitem{ashfaq2014spatial}
A~Ashfaq, M~Ashraf, and A~Bahzad.
\newblock Spatial and temporal assessment of groundwater behaviour in the soan
  basin of pakistan.
\newblock {\em University of Engineering and Technology Taxila. Technical
  Journal}, 19(1):12, 2014.

\bibitem{shahid2018understanding}
Muhammad Shahid, Zhentao Cong, and Danwu Zhang.
\newblock Understanding the impacts of climate change and human activities on
  streamflow: a case study of the soan river basin, pakistan.
\newblock {\em Theoretical and Applied Climatology}, 134(1-2):205--219, 2018.

\bibitem{mwangi2016relative}
Hosea~M Mwangi, Stefan Julich, Sopan~D Patil, Morag~A McDonald, and Karl-Heinz
  Feger.
\newblock Relative contribution of land use change and climate variability on
  discharge of upper mara river, kenya.
\newblock {\em Journal of Hydrology: Regional Studies}, 5:244--260, 2016.

\bibitem{ficchi2014short}
Andrea Ficchi, Luciano Raso, Pierre-Olivier Malaterre, David Dorchies, Maxime
  Jay-Allemand, Francesca Pianosi, Peter-Jules van Overloop, and Guillaume
  Thirel.
\newblock Short term reservoirs operation on the seine river: Performance
  analysis of tree-based model predictive control.
\newblock 2014.

\bibitem{rehman2018optimization}
Sadaqat Rehman, Shanshan Tu, Obaid Rehman, Yongfeng Huang, Chathura
  Magurawalage, Chin-Chen Chang, et~al.
\newblock Optimization of cnn through novel training strategy for visual
  classification problems.
\newblock {\em Entropy}, 20(4):290, 2018.

\bibitem{tu2017csfl}
Sadaqat~ur Rehman, Shanshan Tu, Yongfeng Huang, Guojie Liu, et~al.
\newblock Csfl: A novel unsupervised convolution neural network approach for
  visual pattern classification.
\newblock {\em AI Communications}, 30(5):311--324, 2017.

\bibitem{boureau2010theoretical}
Y-Lan Boureau, Jean Ponce, and Yann LeCun.
\newblock A theoretical analysis of feature pooling in visual recognition.
\newblock In {\em Proceedings of the 27th international conference on machine
  learning (ICML-10)}, pages 111--118, 2010.

\bibitem{hochreiter1998vanishing}
Sepp Hochreiter.
\newblock The vanishing gradient problem during learning recurrent neural nets
  and problem solutions.
\newblock {\em International Journal of Uncertainty, Fuzziness and
  Knowledge-Based Systems}, 6(02):107--116, 1998.

\bibitem{Hochreiter1997Long}
Sepp Hochreiter and J{\"u}rgen Schmidhuber.
\newblock Long short-term memory.
\newblock {\em Neural computation}, 9(8):1735--1780, 1997.

\bibitem{Zeiler2013Visualizing}
Matthew~D Zeiler and Rob Fergus.
\newblock Visualizing and understanding convolutional networks (2013).
\newblock {\em arXiv preprint arXiv:1311.2901}, 2013.

\bibitem{suykens1999least}
Johan~AK Suykens and Joos Vandewalle.
\newblock Least squares support vector machine classifiers.
\newblock {\em Neural processing letters}, 9(3):293--300, 1999.

\bibitem{harrell2015ordinal}
Frank~E Harrell.
\newblock Ordinal logistic regression.
\newblock In {\em Regression modeling strategies}, pages 311--325. Springer,
  2015.

\bibitem{friedman2002stochastic}
Jerome~H Friedman.
\newblock Stochastic gradient boosting.
\newblock {\em Computational Statistics \& Data Analysis}, 38(4):367--378,
  2002.

\bibitem{liaw2002classification}
Andy Liaw, Matthew Wiener, et~al.
\newblock Classification and regression by randomforest.
\newblock {\em R news}, 2(3):18--22, 2002.

\bibitem{nateghi2011comparison}
Roshanak Nateghi, Seth~D Guikema, and Steven~M Quiring.
\newblock Comparison and validation of statistical methods for predicting power
  outage durations in the event of hurricanes.
\newblock {\em Risk Analysis: An International Journal}, 31(12):1897--1906,
  2011.

\bibitem{nateghi2016statistical}
Roshanak Nateghi, Jeremy~D Bricker, Seth~D Guikema, and Akane Bessho.
\newblock Statistical analysis of the effectiveness of seawalls and coastal
  forests in mitigating tsunami impacts in iwate and miyagi prefectures.
\newblock {\em PloS one}, 11(8):e0158375, 2016.

\bibitem{grizzetti2017human}
B~Grizzetti, A~Pistocchi, C~Liquete, A~Udias, F~Bouraoui, and W~Van De~Bund.
\newblock Human pressures and ecological status of european rivers.
\newblock {\em Scientific reports}, 7(1):205, 2017.

\bibitem{nishina2017varying}
Kazuya Nishina, Mirai Watanabe, Masami~K Koshikawa, Takejiro Takamatsu,
  Yu~Morino, Tatsuya Nagashima, Kunika Soma, and Seiji Hayashi.
\newblock Varying sensitivity of mountainous streamwater base-flow no 3-
  concentrations to n deposition in the northern suburbs of tokyo.
\newblock {\em Scientific reports}, 7(1):7701, 2017.

\bibitem{lancini1997diagnosing}
Stefano Lancini, Marco Lazzari, Alberto Masera, and Paolo Salvaneschi.
\newblock Diagnosing ancient monuments with expert software.
\newblock {\em Structural engineering international}, 7(4):288--291, 1997.

\end{thebibliography}

\end{document}